# Tax Knowledge Graph for a Smarter and More Personalized TurboTax

Jay Yu
Technology Futures
Intuit
San Diego, California, USA
jay_yu@intuit.com

Kevin McCluskey
Technology Futures
Intuit
San Diego, California, USA
kevin_mccluskey@intuit.com

Saikat Mukherjee
Technology Futures
Intuit
Mountain View, California, USA
saikat_mukherjee@intuit.com

## ABSTRACT

Most knowledge graph use cases are data-centric, focusing on representing data entities and their semantic relationships. There are no published success stories to represent large-scale complicated business logic with knowledge graph technologies.

In this paper, we will share our innovative and practical approach to representing complicated U.S. and Canadian income tax compliance logic (calculations and rules) via a large-scale knowledge graph. We will cover how the Tax Knowledge Graph is constructed and automated, how it is used to calculate tax refunds, reasoned to find missing info, and navigated to explain the calculated results.

The Tax Knowledge Graph has helped transform Intuit's flagship TurboTax product into a smart and personalized experience, accelerating and automating the tax preparation process while instilling confidence for millions of customers.

## 1 Introduction

Intuit, the maker of the market-leading TurboTax product, is in a multi-year transformational journey to become an AI-driven Expert Platform company. In this new AI era, robustness and trustworthiness of the tax calculation engine, which is the heart of TurboTax, have become table stakes and are no longer the differentiating customer delighters. Customers have higher expectations for an intelligent and personalized experience to get more money, with minimal work and complete confidence. Below we will share the traditional approach of tax software to tackle the complexity of U.S. tax logic and explain why it is no longer adequate to meet our customers' needs.

### 1.1 Complexity of U.S. Tax Laws

According to [1] and [2], the level of complexity in the U.S. Tax Code kept increasing dramatically over time, which is reflected by the following statistics:

- 74,608 pages describing the tax code (by 2014 [1])
- Close to 800 federal tax forms, plus additional forms from 45 states
- Yearly changes, often with last-minute changes driven by late tax legislations

This complexity makes it very hard for many of the 150 million U.S. taxpayers, who spend an estimated total of 7 billion hours every year on tax preparation, to file their taxes. They struggle to ensure their tax returns are done accurately and with confidence. That's why TurboTax has been able to offer tremendous value to greatly alleviate the burden of filing a tax return with guaranteed accuracy and ease of use.

### 1.2 Traditional Approach to Codify Tax Logic

The conventional way to codify tax logic is to use a procedural programming language to hard code all the logic per tax form instruction with consultation from tax domain experts. Below is a simple example from part of the IRS 1040 tax form and the associated conventional programming logic.

**Figure 1: Section of U.S. IRS Individual Tax Form 1040**

```
L17 = getTaxWithheld() # from W-2 and 1099

L19 = L17 + L18e # total payments
if L19 > L16:
    L20 = L19 - L16 # refund
else:
    L20 = 0
```

**Figure 2: Sample Procedural Calculation Code**

While the sample code in Figure 2 may seem straightforward from this simple example, there are a number of key limitations:

- Procedural programming - done by **programmers** who rely on tax domain experts for the logic spec
- Top-down sequential execution - have to **re-calculate the entire return** with one input change
- All inputs required for complete calc - need to **collect all information upfront**
- Implicit explainability hidden in code - cost prohibitive to open the calc black box

When the above code is scaled to hundreds of thousands of calculations and rules for the entire tax return, these limitations become *huge barriers* to our goal of providing a highly personalized experience to get taxes done right with minimal effort and high confidence. This is where we leverage Tax Knowledge Graph for a *fundamental paradigm shift* in our approach to represent complicated tax compliance calculations and rules at scale and connect associated user data together.

## 2 Tax Knowledge Graph

The definition of knowledge graph varies from academic research ([3]) to practical implementation approaches and target use cases ([4]). We created an open and inclusive version of the definition below to make it easier to understand and applicable to our domain.

"Knowledge graph is a graph to **represent data** (facts, information) **and logic** (conditions) into concepts as vertices, and relationships between concepts as edges, **created / enriched** from experience and learning, used for **understanding and reasoning**, by **human and machine**."

We took a **logic-centric** approach to apply knowledge graph technology to the tax compliance domain. Our Tax Knowledge Graph is designed to represent tax calculation logic and rules with two major components:

1. **Calc Graph**: a comprehensive graph with tens of thousands calculation statements represented as interconnected calc modules, with each as a calc function node connecting input and output data nodes together.

2. **Completeness Graph**: a special type of graph to represent complicated decision tree logic to determine applicability of specific tax topics, such as earned income credit, based on user data.

We will cover the detail for both below.

### 2.1 Calc Graph

The Calc Graph consists of the meta-level *Generic Calc Patterns (GISTs)* curated by engineers and tax experts to model generic calc patterns, and the detailed *Bounded Calc Models (GIST instances)* developed by tax domain experts in the context of a tax form by applying matching GISTs to a specific instruction. Below is an example Calc Graph construction for the simple example from IRS 1040 tax form in Figure 1.

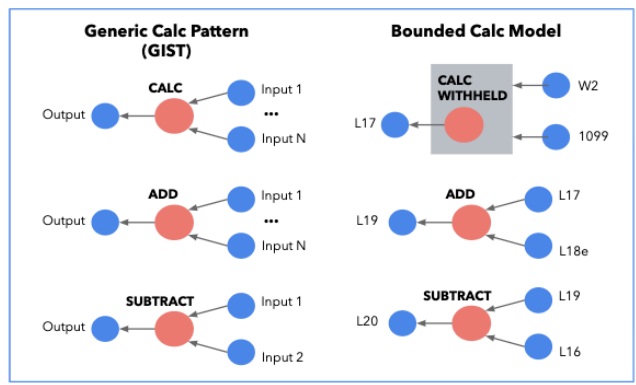

**Figure 3: Sample GISTs and Components in Calc Graph**

In this example, there are 3 GISTs identified in Figure 3:

- CALC: a generic function taking many input values and producing an output value
- ADD: adding all input values and producing the single value as the sum of all input values
- SUBTRACT: calculating the difference between two input values

These GISTs are used by tax domain experts to define calculation relationships among lines on the tax forms, by mapping a specific line instruction (e.g., L20) to a GIST (SUBTRACT), and binding the input values and output value from the GIST to the form lines. The resulting GIST instance becomes a derived "Bounded Calc Model" for the GIST within the context of the current tax form.

By chaining all these "Bounded Models" together, we form a comprehensive Calc Graph in Figure 4 with the following benefits:

- Declarative programming makes it **easy for tax domain experts** (non-programmers).
- Granular, incremental composition allows **decomposition of complicated tax logic.**

- Visible calc dependency and data flow drives **execution efficiency and testability.**
- Built-in, explicit explainability enables the user experience to **open the calc black box.**

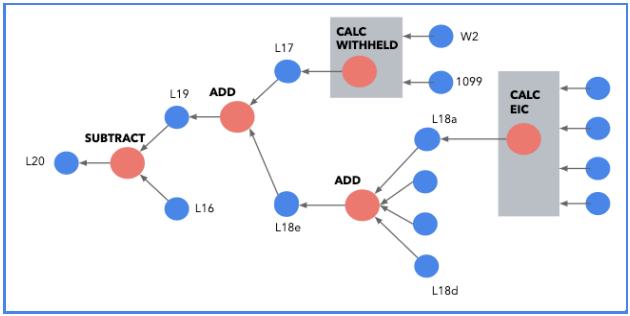

**Figure 4: Chaining Components for Calc Graph**

The Calc Graph is our knowledge graph for the tax domain calculation logic, which is the input to our runtime Tax Knowledge Engine that we will cover later in section 3 of this paper.

## 2.2 Built-in Explainability

With calculation logic represented as a knowledge graph, we get immediate visibility and insights into the relationship of various calculations via their connected input and output variables. Thus, it is much easier to understand the outcome of the calculation by analyzing and traversing the graph.

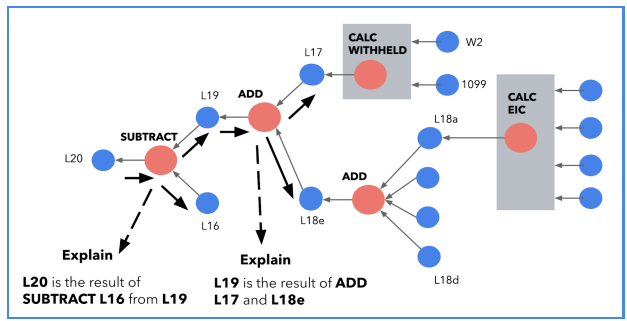

**Figure 5: Explanation via Backward Chaining**

Figure 5 shows how the Calc Graph from Figure 4 can be used to explain the outcome of calculation for L20. This can be done by backward chaining the Calc Graph. First, find the GIST instance that produces L20 and its associated inputs (L19 and L16), and generate a user-friendly explanation of the GIST instance. Then drill down to the next level by traversing backward through the input values, driven by user request or automation rules. This explainability has been packaged into the TurboTax product feature called **_Explain Why_** with great results to drive user delight and confidence.

## 2.3 Completeness Graph

While Calc Graphs are great at modeling calculations with built-in explainability, there are many complicated eligibility rules (e.g., qualification for child tax credit) that are not the best fit to the calculation pattern. To simplify the definition and processing of such logic, we created a special Completeness Graph to break down the complexity and minimize user efforts to determine eligibility. Below is a simple example to determine if somebody is qualified for a tax benefit:

- If a person is not a resident of California, he/she is not qualified;
- Otherwise, he/she must be older than 18 to qualify for the benefit.

These eligibility rules usually include a starting point, set of conditions and its associated input variables, and a finite number of outputs to provide the answer. We model this class of logic with a Completeness Graph backed by a truth table, which consists of a starting node, a number of output nodes for decisions, input variable nodes and conditional nodes. See Figure 6 for the Completeness Graph representation for the above example.

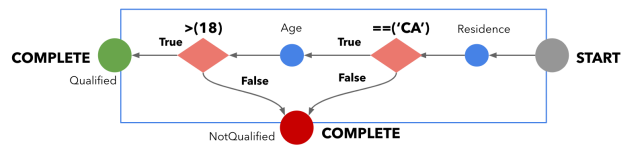

| Residence == 'CA' | Age > 18 | Outcome |
|---|---|---|
| True | True | Qualified |
| False | True | Disqualified |
| True | False | Disqualified |
| False | False | Disqualified |

**Figure 6: Completeness Graph and Decision Table**

With the Completeness Graph, we can rely on back chaining (from COMPLETE nodes) to determine what information is missing in order to complete the eligibility test. In this example, if we already know age is 17, we can reason with the completeness graph and associated truth table to find out whether residence info is necessary to reach to the final decision.

Completeness Graph is the key enabler for a dynamic **data-driven, personalized experience** that **_minimizes questions_** for data collection, since it can constantly prune down the alternative paths and remove unnecessary questions from the user experience. Therefore, it played a key role for TurboTax to deliver on the promise of taxes done with minimal or no work and high confidence.

## 3 Tax Knowledge Engine

With the Tax Knowledge Graph defined to represent the calculations and rules behind the entire tax return, we build a runtime engine to instantiate an instance of the Tax Knowledge Graph for a taxpayer's use data, in order to make it actionable in TurboTax. Tax Knowledge Engine is designed to integrate all the key components together: user data, Tax Knowledge Graph and TurboTax UI application and services.

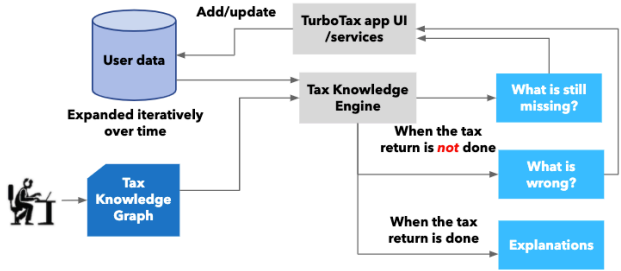

**Figure 7: Tax Knowledge Engine – E2E Architecture**

Figure 7 shows the end-to-end architecture of how the Tax Knowledge Engine combines the instantiated instance of the Tax Knowledge Graph with a particular user's data, to orchestrate a personalized and dynamic user experience that can show user what's ***missing***, what's ***wrong***, at ***any moment*** and for ***any (partial) user data***, and ***explain*** back how it's done.

It starts with tax domain experts to define tax calculation logic via constructing the Tax Knowledge Graph. To reduce the complexity, we chose to use a simple XML to store the definition of the Calc and Completeness Graphs and created a set of ease-to-use domain-specific tools for tax developers without the need to deal with knowledge graph technologies directly (e.g., RDF, SPARQL …).

The Tax Knowledge Engine is implemented in C++ to load and parse Calc and Completeness Graphs in memory, and to use them to drive the user data instantiation, calculation processing, completeness reasoning and explanation generation. The engine will inform TurboTax application about the latest calculation result, whether the return is complete, and any errors found in the tax return. In addition, it will also provide explanation per UI experience flow.

When TurboTax application collects more user inputs, it will flow them back to the Tax Knowledge Engine to determine which subset of the knowledge graphs need to be updated/re-calculated before providing the updated results back.

## 3.1 Calc Graph Automation

Even though the new Tax Knowledge Graph approach has many advantages over the traditional tax calc programming approach, it still requires manual efforts to build and update the graph every year, where tax analysts read forms and instructions, understand the calc to be performed, and translate into artifacts in the knowledge graph. This process is often error-prone and becomes increasingly difficult to scale horizontally if we have to train new tax analysts to learn the tools and process. By using natural language processing and machine learning technologies, we have developed techniques that successfully extract calculations from agency content with minimal human supervision. This is accomplished via the following automated steps:

1. Extract content from the PDF version of tax forms and instructions and convert it into a machine processable simplified representation.
2. Identify terms that represent atomic concepts within our tax domain.
3. Perform semantic pattern matching to identify the operations and operands outlined in the forms.
4. Generate calculations in Tax Knowledge Graph format from matching GISTs and data model.

For extremely complicated calculations, tools are created to have experts (human-in-loop) to help break them down into smaller and less complicated modules for automation.

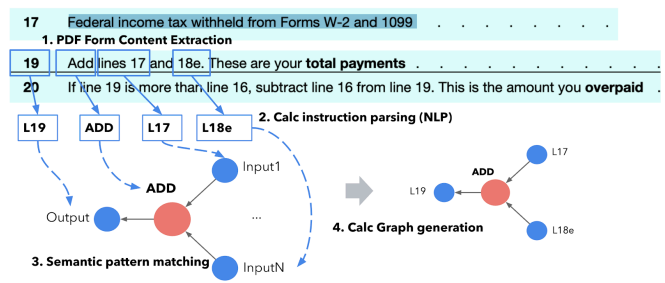

**Figure 8: Automating Calc Graph Construction**

Figure 8 is a simple example to illustrate the process to automate the Calc Graph construction in the following steps:

1. Extract the content of Line 19 from the original PDF document.
2. Parse instruction for line 19 via NLP and extract the following terms: L19 as output tax field, Add as calc, L17 and L18e as input tax fields.
3. Perform a semantic pattern search on the GIST library (calc patterns) with the terms from Step 2, which finds the ADD GIST.
4. Bound input and output fields from the terms to the ADD GIST, resulting in a concrete calc statement with an instantiated GIST bound by tax fields in the current context of a tax form.

The PDF extractor developed in Step 1 identifies various meta-data in these forms such as form sections, lines with their descriptions, fields and tables. It also extracts constants and connections across lines between forms. We are able to automatically create a knowledge graph over the U.S. tax domain, consisting of fields in forms and connections between them, with over ***987 forms*** and ***25,729 connections*** between them.

The term extraction technique [7] developed in Step 2 helped us identify close to **13,000 terms** by parsing 82,900 sentences from tax forms and 187,600 sentences from corresponding instructions from the IRS, which are used for automated Calc Graph generation.

Finally, the automated calc graph generation algorithm under Step 3 and 4 above was able to automate **72%** of the Canadian income tax form fields with calculation instructions, covering **5,496 fields** out of the 7,590 calculation fields.

## 4  Results

We are able to deliver game-changing capabilities via the Tax Knowledge Graph to make TurboTax much smarter and more personalized based on the Tax Knowledge Graph. Below is a recent example.

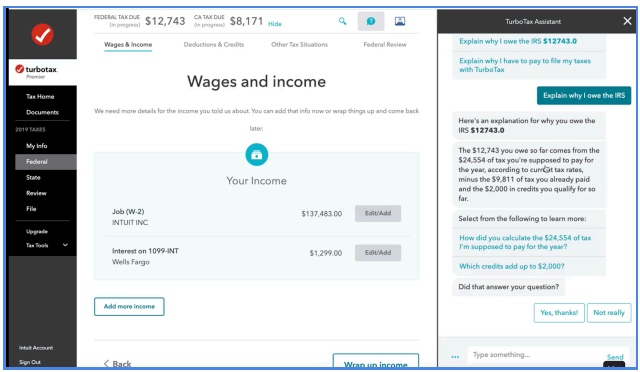

**Figure 9: Explain Why in TurboTax Assistant**

Figure 9 shows the new Explain Why feature integrated inside the TurboTax Assistant Chat Bot experience (on the right pane). This interactive and natural language-based explanation capability is context-aware, deeply integrated with a user's tax return data, and highly personalized to his/her tax scenario. It allows a user to dynamically drill down to any level of detail on how the tax refund or amount owed is calculated, without the need to contact a tax professional.

Our customers benefit directly from these innovations powered by Tax Knowledge Graph in **speed to refund**, with **high confidence** for peace of mind. During the initial limited deployment in the first year (2015), we saw a **46% reduction in support calls** on one of the top call driver topics (W2), and a **77% helpful rating** for the Explain Why feature from 16M customers.

Our employees also benefit directly from the Tax Knowledge Graph in **productivity, quality and scale** of tax logic development. This is accomplished via the simplified **development experience for tax domain experts** to specify, update, test and deploy tax logic to drive data-driven dynamic user experiences at scale. The automated Calc Graph generation helped with a **7x speed-up** in creating the knowledge graphs for Canadian income tax calculations compared to a manual approach.

Combining the above customer and employee benefits, Tax Knowledge Graph has played a major role in Intuit Consumer Group's **year-over-year double-digit revenue growth** in the highly competitive tax preparation software marketplace.

## 5  Related Work

Most popular large-scale knowledge graph applications center around data [5]: e.g., Information Graph from Google, Professional Graph from LinkedIn (Microsoft), Social Graph from Facebook, Product Graph from eBay. Our Tax Knowledge Graph takes a logic-centric approach with focus on compliance calculations and rules. It is the first large-scale knowledge graph we know of to cover tax compliance behind the entire U.S. Federal Tax Code.

There are growing interests in applying knowledge graph and NLP to document understanding in the tax compliance domain, extracting semantic relationships between concepts from compliance forms and instructions, in order to better answer user questions. One recent example is the attempt to use ontology to determine legal implication of corporate transactions [6]. Our focus is to automate the calculation and rules generation from IRS forms.

Recent work from Intuit [7] shared how we use NLP technologies to detect and label executable calculations from IRS tax forms. It is one of the key components to automate the Tax Knowledge Graph construction and maintenance. To the best of our knowledge, there is no published research or industry work focusing on calculation logic.

## 6  Summary

In this paper, we described a unique and innovative way to represent the complicated logic in the U.S. income tax domain via a large-scale Tax Knowledge Graph. This has transformed how we build and maintain tax logic with a combination of tools automation and human in the loop. The Tax Knowledge Graph has played a significant role in making Intuit's flagship TurboTax product smarter and more personalized, so customers can get their taxes done right with minimal effort and high confidence.

We believe that we have just scratched the surface of logic-centric knowledge graph technology as our initial focus on the tax domain. We believe such an approach can be applicable to other domains. We have started working on a more generic Knowledge Engine Platform to scale this technology for many more use cases outside the tax domain.

### ACKNOWLEDGMENTS

We would like to thank the following people for their significant contribution to the Tax Knowledge Graph and related technology innovation at Intuit: Gang Wang, Alex Balazs, Mike Artamonov,

Jennifer Keenan, Rushabh Mehta, Saneesh Joseph, Cem Unsal, Dave HaneKamp, Marty Lewandowski, Matt Brincho, Steven Atkinson, Sudhir Agarwal, Esme Manandise, Anu Singh, Mritunjay Kumar, Conrad DePeuter, Karpaga Ganesh Patchirajan, Anu Sreepathy, Clarence Huang, Yi Ng, Cindy Osmon, Bharath Kadada, and many other engineers, tax experts and product managers/leaders in Intuit U.S. and Canada who helped refine and implement the ideas and bring them into the TurboTax product lines.

We would also like to thank Karen Weiss for the editorial and formatting help.

In addition, the key innovations described in this paper are covered by the following granted U.S. patents: 7,765,097, 10,169,826, 10,387,969, 10,387,970, 10,402,913, 10,475,132, 10,475,133, 10,579,721, 10,664,925, 10,664,926.

## REFERENCES

[1] Jason Russell. 2016. Look at how many pages are in the federal tax code. *Washington Examiner*: https://www.washingtonexaminer.com/look-at-how-many-pages-are-in-the-federal-tax-code

[2] Walsh, Robert J. 2019. A History of U.S. Tax Code Complexity within Computer-Based Return Preparation. *Accounting & Taxation*, v. 11 (1) p. 47-57: https://ssrn.com/abstract=3480705

[3] Definition of Ontology (2020). Wikipedia: https://en.wikipedia.org/wiki/Ontology_(information_science)

[4] Definition of Knowledge Graph (2020). Wikipedia: https://en.wikipedia.org/wiki/Knowledge_Graph

[5] Yuqing Gao, Anant Narayanan, Alan Patterson, Jamie Taylor, Anshu Jain. 2018. Panel on Enterprise-Scale Knowledge Graph. *International Conference on Semantic Web (ICSW 2018)*: http://iswc2018.semanticweb.org/panel-enterprise-scale-knowledge-graphs/index.html

[6] Yoo Jung An, Ned Wilson. 2016. Tax Knowledge Adventure: Ontologies that Analyze Corporate Tax Transactions. *dg.o '16*, June 08-10, 2016, Shanghai, China: https://dl.acm.org/doi/pdf/10.1145/2912160.2912200

[7] Esme Manadise. 2019. Towards Unlocking the Narrative of the United States Income Tax Forms. *Proceedings of the Second Financial Narrative Processing Workshop (FNP 2019)*, September 30, Turku Finland: https://www.ep.liu.se/en/conference-article.aspx?series=ecp&issue=165&Article_No=5